\documentclass[review]{elsarticle}

\usepackage{graphicx}
\usepackage{indentfirst}
\usepackage{amsfonts,amssymb}
\usepackage{multirow}
\usepackage{float}
\usepackage{threeparttable}
\usepackage{enumerate}
\usepackage{geometry}
\usepackage{booktabs}
\usepackage{longtable}
\usepackage{lscape}
\usepackage{bm}
\usepackage{hyperref}       
\usepackage{graphicx}
\usepackage{amsmath}
\usepackage{amssymb}
\usepackage[ruled,vlined,linesnumbered,titlenumbered]{algorithm2e}
\usepackage{lscape}
\usepackage{subfigure}
\usepackage{color}

\usepackage{lineno,hyperref}
\modulolinenumbers[5]

\journal{arXiv}

\begin{document}









\thispagestyle{plain}
\begin{frontmatter}
\title{Weighting and Pruning based Ensemble Deep Random Vector Functional Link Network for Tabular Data Classification}



\author[rvt]{Qiushi Shi}
\ead{qiushi001@e.ntu.edu.sg}
\author[rvt]{Ponnuthurai Nagaratnam Suganthan\corref{cor1}}
\ead{epnsugan@ntu.edu.sg}
\author[rvt]{Rakesh Katuwal}
\ead{rakeshku001@e.ntu.edu.sg}

\address[rvt]{School of Electrical and Electronic Engineering, Nanyang Technological University, 50 Nanyang Avenue, Singapore 639798}
\begin{abstract}
In this paper, we first introduce batch normalization to the edRVFL network. This re-normalization method can help the network avoid divergence of the hidden features. Then we propose novel variants of Ensemble Deep Random Vector Functional Link (edRVFL). Weighted edRVFL (WedRVFL) uses weighting methods to give training samples different weights in different layers according to how the samples were classified confidently in the previous layer thereby increasing the ensemble's diversity and accuracy. Furthermore, a pruning-based edRVFL (PedRVFL) has also been proposed. We prune some inferior neurons based on their importance for classification before generating the next hidden layer. Through this method, we ensure that the randomly generated inferior features will not propagate to deeper layers. Subsequently, the combination of weighting and pruning, called Weighting and Pruning based Ensemble Deep Random Vector Functional Link Network (WPedRVFL), is proposed. We compare their performances with other state-of-the-art deep feedforward neural networks (FNNs) on 24 tabular UCI classification datasets. The experimental results illustrate the superior performance of our proposed methods.    
\end{abstract}
\end{frontmatter}

\section{Introduction}
\indent 
Deep learning has been extremely successful in recent years. Ranging from vision and video tasks to natural language processing, these deep neural networks have reached state-of-the-art results in multiple domains \citep{lecun2015deep,schmidhuber2015deep}. In conventional neural networks, back-propagation methods are used to train a large number of parameters in these models \citep{plaut1987learning}. Although such a training method makes it possible to optimize the parameters, the time-consuming training process has become a severe problem in recently designed complex neural networks. Also, a BP-trained neural network may fall into a local minimum and gives a sub-optimal result \citep{bengio2013deep,suganthan2018non,suganthan2021origins}. By looking at the Kaggle competitions that have no relation with vision or sequence, we can easily find that deep learning is not always the best solution for diverse tasks \citep{olson2018modern,shavitt2018regularization}.

At the same time, another kind of neural network based on randomization is attracting significant attention because of its superiority to overcome the shortcomings of the conventional models \citep{schmidt1992feed,suganthan2018non,widrow2013no}. It has been successfully applied to a range of tasks from classiﬁcation \citep{giryes2016deep,junior2018randomized,zhang2016comprehensive}, regression \citep{he2018random,vukovic2018comprehensive}, visual tracking \citep{zhang2016visual}, to forecasting \citep{ren2016random,wang2017forecasting}. Instead of using back-propagation to train, this randomization-based neural network frequently uses a closed-form solution to optimize parameters in the model \citep{te1995random}. Unlike the BP-trained neural networks which need multiple iterations, the randomization-based neural networks only need to be trained once by feeding all the samples to the model together. Among these models, Random Vector Functional Link Network (RVFL) \citep{pao1992functional} is a typical representative with a single hidden layer. Its universal approximation ability has been proved in \citep{needell2020random}. The weights and biases are randomly generated in this neural network. And its uniqueness lies in a direct link that connects the information from the input layer to the output layer. However, due to different random seeds and perturbations in the training set, this randomized neural network can perform quite differently in each realization~\citep{li2017insights}. To increase the performance, stability, and robustness of this model, two improved structures named Deep Random Vector Functional Link Network (dRVFL) and Ensemble Deep Random Vector Functional Link Network (edRVFL) were proposed~\citep{Shi2021Random}. The dRVFL network is a deep version of RVFL network, which allows the existence of multiple hidden layers, while edRVFL network treats each hidden layer as a classifier to compose an ensemble.
\par
However, with the edRVFL network goes deeper, the divergence of the randomized hidden features will become a serious problem. Therefore, using normalization methods to re-normalize the hidden features is extremely important for improving the performance of the edRVFL network. In this paper, we employ the batch normalization scheme \citep{ioffe2015batch} to do the re-normalization work. To the best of our knowledge, this is the first time that batch normalization is introduced to the randomized neural network. After the re-normalization process, the mean and the variance of the hidden features will become 0 and 1. Then, we scale and shift these values to increase the expression capacity of the neural network.
\par
Besides, there are still some drawbacks to the edRVFL network. Firstly, for every layer (or classifier) in the edRVFL network, they share the same training samples. Meanwhile, these training samples have the same weights in the training process. Compared to ensemble methods that using differing training bags for each classifier, these ensemble frameworks which utilize similar training sets usually perform worse ~\citep{liaw2002classification,rodriguez2006rotation}. Moreover, the testing accuracy for the last few layers may slightly go down when the network becomes deeper. We believe that some inferior features can be generated since we randomly generate the weights for the hidden neurons. And these useless features will propagate to deeper layers inducing further inferior features to decrease the overall testing accuracy. 
\par
Thus, for solving the first problem, we introduce a weighting matrix. Each training sample will be allocated a particular weight when performing the closed-form solution depending on its performance in the previous layer. Our approach differs from Weighted Extreme Learning Machine \citep{zong2013weighted} which gives weights to each sample for addressing the problem of imbalance learning. The main purpose is to ensure that different classifiers can have their preference for a particular portion of the training samples that were not classified with high confidence in the previous layer. We have also tried to apply the sample weighting method of Adaboost \citep{freund1996experiments}. However, most of the samples will be given weights near zero while only a few can be allocated reasonable weights. Therefore, we propose four different weighting methods in this paper, and this improved variant of edRVFL network is named Weighted Ensemble Deep Random Vector Functional Link Network (WedRVFL).
\par
Besides, pruning algorithms are widely used to reduce the heavy computational cost of deep neural networks in low-resource settings \citep{liu2018rethinking}. Different effective techniques have been proposed to cut off the redundant part of the neural network models \citep{lecun1989optimal,hassibi1993second,han2015learning,molchanov2016pruning}. In our case, we perform it by selecting some inferior features in the hidden layer and prune them permanently. The selection process can help to prevent the propagation of inferior features and maintain the testing accuracy for deeper layers. We named this improved variant of edRVFL network as Pruning-based Ensemble Deep Random Vector Functional Link Network (PedRVFL). Although there was previous work that applying pruning strategy to the RVFL network in \citep{henriquez2018non}, we would like to highlight that our work is different from theirs at the following point: They do pruning after training to shrink the size of the neural network. However, we perform pruning during the generation step so that inferior features will not propagate to deeper layers. Additionally, we integrate the advantages of WedRVFL and PedRVFL to create a combined model called Weighting and Pruning based Ensemble Deep Random Vector Functional Link Network (WPedRVFL).
\par
The key contributions of this paper are summarized as follows:
\begin{itemize}
    \item We introduce the batch normalization to the edRVFL network for re-normalizing the hidden features.
    \item We employ the weighting scheme to allocate different weights to different samples in the edRVFL network. We name it WedRVFL network. The weight matrix changes according to the samples' predictions in the previous layers. This method can make sure that each hidden layer in the network has different biases for each sample and increase the ensemble classification accuracy.
    \item We propose pruning based edRVFL network called PedRVFL network. Instead of pruning neurons after the training process, we cut off the inferior neurons according to their importance for classification when we are training the model. This method can prevent the propagation of detrimental features and increase the classification accuracy in deeper layers.
    \item The combination of weighting and pruning based edRVFL network named WPedRVFL network is also presented in the paper.
    \item The empirical results show the superiority of our new methods over 11 state-of-the-art methods on 24 UCI benchmark datasets.
\end{itemize}

The rest of the paper is organized as follows: Section 2 outlines the basic concepts of RVFL network and illustrates the ensemble deep version of this structure. Section 3 introduces the re-normalization method for the edRVFL network. Then Section 4 gives details about our new proposed versions of edRVFL network. In Section 5, the performance of our methods, as well as other deep feedforward neural networks (FNNs) and RVFL variants are compared. Finally, conclusions and future research directions are presented in Section 6.
\section{Related works}
\indent 
In this section, we give a brief review of the structure of the standard RVFL network and ensemble deep RVFL network.
\subsection{Random Vector Functional Link Network}
\begin{figure}[h!]
\centering
\includegraphics[scale=0.4]{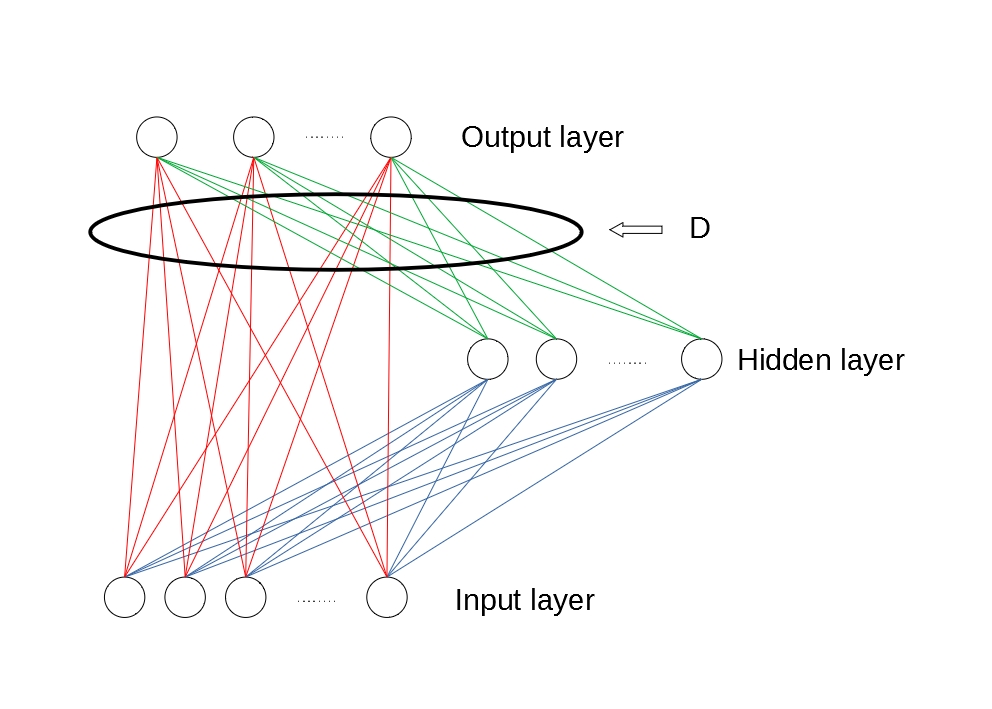}
\caption{The structure of RVFL network. The original features have two ways to transfer to the output layer: One is going through the hidden layer(the blue lines and the green lines), and the other one is transferred through the direct link(the red lines).}
\label{fig:fig1}
\end{figure}

\indent
As shown in Fig.\ref{fig:fig1}, a basic RVFL network consists of one input layer, one hidden layer, and one output layer \citep{pao1992functional}. Unlike the general neural network, the uniqueness of RVFL network is its direct link between the input layer and the output layer. This framework conveys both the linear features in the input layer and the non-linearly transformed features in the hidden layer to the output layer. Because the parameters for the hidden layer are randomly generated and kept fixed during the training process, the only thing it needs to learn is the output weights $\bm{\beta}$. The solution of the $\bm{\beta}$ can be computed by solving the optimization problem given as follows:
\begin{equation}
O_{RVFL}=\mathop{\min}\limits_{\bm{\beta}}||\mathbf{D}\bm{\beta}-\mathbf{Y}||^2_2+\lambda||\bm{\beta}||^2_2
\end{equation}
where $\mathbf{D}$ represents all the input features and output features, $\mathbf{Y}$ is the true vector we want to fit, and $\lambda$ is a regularization parameter that controls how much the RVFL network cares about its model complexity. 
\par
Normally, this kind of optimization problem can be solved via Moore-Penrose pseudoinverse \citep{barata2012moore} and ridge regression \citep{hoerl1970ridge}. For Moore-Penrose pseudoinverse, the algorithm does not consider the contribution of the part $||\bm{\beta}||^2$, and $\lambda$ is simply set to 0. So the solution is given by:
\begin{equation}
\bm{\beta}=\mathbf{D^+Y}
\end{equation}
Moreover, for the ridge regression where $\lambda$ is not equal to 0, the solution can be written as:
\begin{equation}
\mbox{Primal Space: } \bm{\beta}=(\mathbf{D^T}\mathbf{D}+{\lambda}\mathbf{I})^{-1}\mathbf{D^T}\mathbf{Y}
\label{equation:primal}
\end{equation}
\begin{equation}
\mbox{Dual Space: } \bm{\beta}=\mathbf{D^T}(\mathbf{DD^T}+{\lambda}\mathbf{I})^{-1}\mathbf{Y}
\label{equation:dual}
\end{equation}
Depending on the number of total feature dimensions, the computational complexity for RVFL training can be reduced by using the primal or dual solution \citep{suganthan2018non}.
\subsection{Ensemble Deep Random Vector Functional Link Network}
\begin{figure}[h!]
\centering
\includegraphics[scale=0.4]{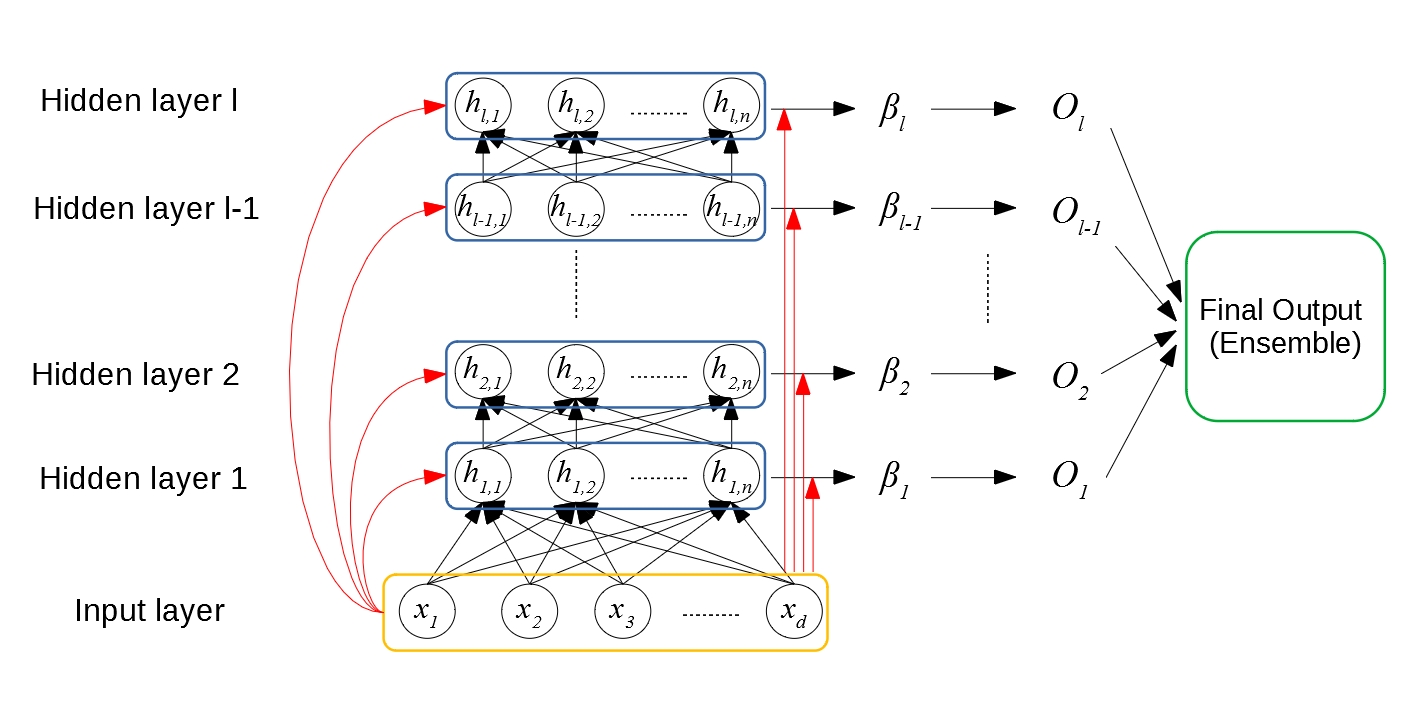}
\caption{The structure of edRVFL network. The whole network can be treated as several independent classifiers, and each classifier $l$ has its own output weights $\beta_l$ and output $O_l$. The final output is obtained by applying some ensemble methods to these independent outputs.}
\label{fig:fig3}
\end{figure}

With deep learning methods become more and more popular today, the deep version of the RVFL networks called Ensemble Deep Random Vector Functional Link Network (edRVFL) was proposed by \citep{Shi2021Random}. It is generally accepted that ensemble learning performs better than a single learner. Inspired by this idea, the author separates the whole network into several independent classifiers. The structure of edRVFL can be found in Fig.\ref{fig:fig3}. 
\par
For easy to understand, here we set the number of hidden neurons in each hidden layer the same to $n$ and the total number of the hidden layers to $l$. Biases will also be omitted in the formula for simplicity. Let $X$ be the input features, and the output of the first hidden layer can be represented as:
\begin{equation}
\mathbf{H}^{(1)}=g(\mathbf{XW}^{(1)}),\quad \mathbf{W}^{(1)} \in\mathbb{R}^{d\times n}
\label{equation:generation1}
\end{equation}
where $d$ is the feature number of the input and $g(\cdot)$ is the non-linear activation function used in each hidden neuron. When $l>1$, this formula becomes:
\begin{equation}
\mathbf{H}^{(l)}=g([\mathbf{H}^{(l-1)}\mathbf{X}]\mathbf{W}^{(l)}),\quad \mathbf{W}^{(l)} \in\mathbb{R}^{(n+d)\times n}
\label{equation:generation2}
\end{equation}

\par
This framework treats every hidden layer as a single classifier. For one classifier (hidden layer), the original features (features in the input layer) and all the hidden features in this layer, are serving as the input for this classifier's prediction. Either Moore-Penrose pseudoinverse \citep{barata2012moore} or ridge regression \citep{hoerl1970ridge} can be chosen for solving the optimization problem. After getting all the hidden layers' predictions, an ensemble method such as major voting or averaging is employed to reach the final output.

\section{Proposed re-normalization scheme for the edRVFL network}

\begin{figure}[h!]
\centering
\includegraphics[scale=0.5]{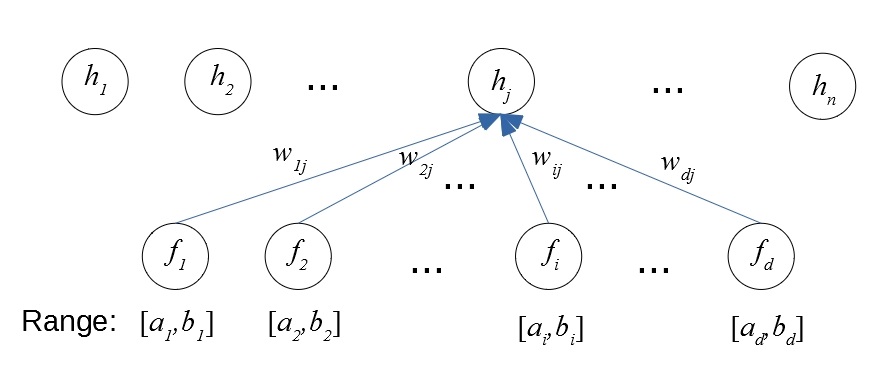}
\caption{The range of the input features and the generation of the first hidden layer.}
\label{fig:divergence}
\end{figure}

For the edRVFL network, the input samples are normalized so that the mean equals to 0 and the variance is 1 for each feature. Also, the hidden weights of the edRVFL network are uniform randomly generated within $[-1,1]$. Suppose the feature number of the training sample is $d$ and let the range of its $i$th normalized feature $f_i$ to be $[a_i,b_i]$, where $a_i\leq0$ and $b_i\geq0$. Then the input of the $j$th node's activation function in the first hidden layer can be written as:

\begin{equation}
    h_j = \sum_{i=1}^{d}w_{ij}f_i
\end{equation}
Where $w_{ij}$ is the hidden weight between the $i$th normalized feature and the $j$th hidden node. We summarize the above operation in Fig. \ref{fig:divergence}.

Based on the condition that $w_{ij}\in[-1,1]$, the range of $h_j$ can be given by $[\sum_{i=1}^{d}a_i,\sum_{i=1}^{d}b_i]$. And this range will keep increasing with more and more hidden layers been generated. Since randomization has the potential to generate inferior features, and this situation can be worse due to this large range. Therefore, re-normalization is crucial for avoiding divergence of the hidden features in the edRVFL network.

Batch normalization is widely used in back-propagation-based deep neural networks to address the Internal Covariate Shift issue during the training process \citep{ioffe2015batch}. In this work, we use it to help the edRVFL network re-normalize its hidden features. To the best of our knowledge, this is the first time that batch normalization is introduced to the randomized neural network. The batch normalization generally has two steps. The first step is to make the inputs' mean and variance to 0 and 1, respectively. Suppose we have $m$ inputs and let the $i$th input to be $x_i$, then the mean value $\mu$ and variance $\sigma^2$ can be obtained by:

\begin{equation}
    \mu = \frac{1}{m}\sum_{i=1}^{m}x_i
\end{equation}

\begin{equation}
    \sigma^2 = \frac{1}{m}\sum_{i=1}^{m}(x_i-\mu)^2
\end{equation}

And the normalization can be done by:
\begin{equation}
    \widehat{x_i} =\frac{x_i-\mu}{\sqrt{\sigma^2+\epsilon}}
\end{equation}

After we have normalized the inputs, we can do scaling and shifting to increase the approximation ability of the network:

\begin{equation}
    y_i = \gamma\widehat{x_i}+\alpha
    \label{equation:re}
\end{equation}

Here $\gamma$ and $\alpha$ are parameters that we need to set for the network. In the back-propagation-based deep neural networks, these two parameters are learned by the back-propagation which are similar to the hidden weights. However, we treat $\gamma$ and $\alpha$ as hyperparameters for the edRVFL network. For simplicity, we keep the $\gamma$ and $\alpha$ fixed for every hidden layer. The validation set is used to help us find the best configuration of these two hyperparameters.

\section{Ensemble Deep Random Vector Functional Link Networks with Weighting $\&$ Pruning }
\indent
In this section, we propose two improved variants of edRVFL network called WedRVFL and PedRVFL in section 4.1 and section 4.2. In section 4.3, their combination WPedRVFL is presented.
\subsection{Weighted Ensemble Deep Random Vector Functional Link Network}

\begin{figure}[h!]
\centering
\includegraphics[scale=0.4]{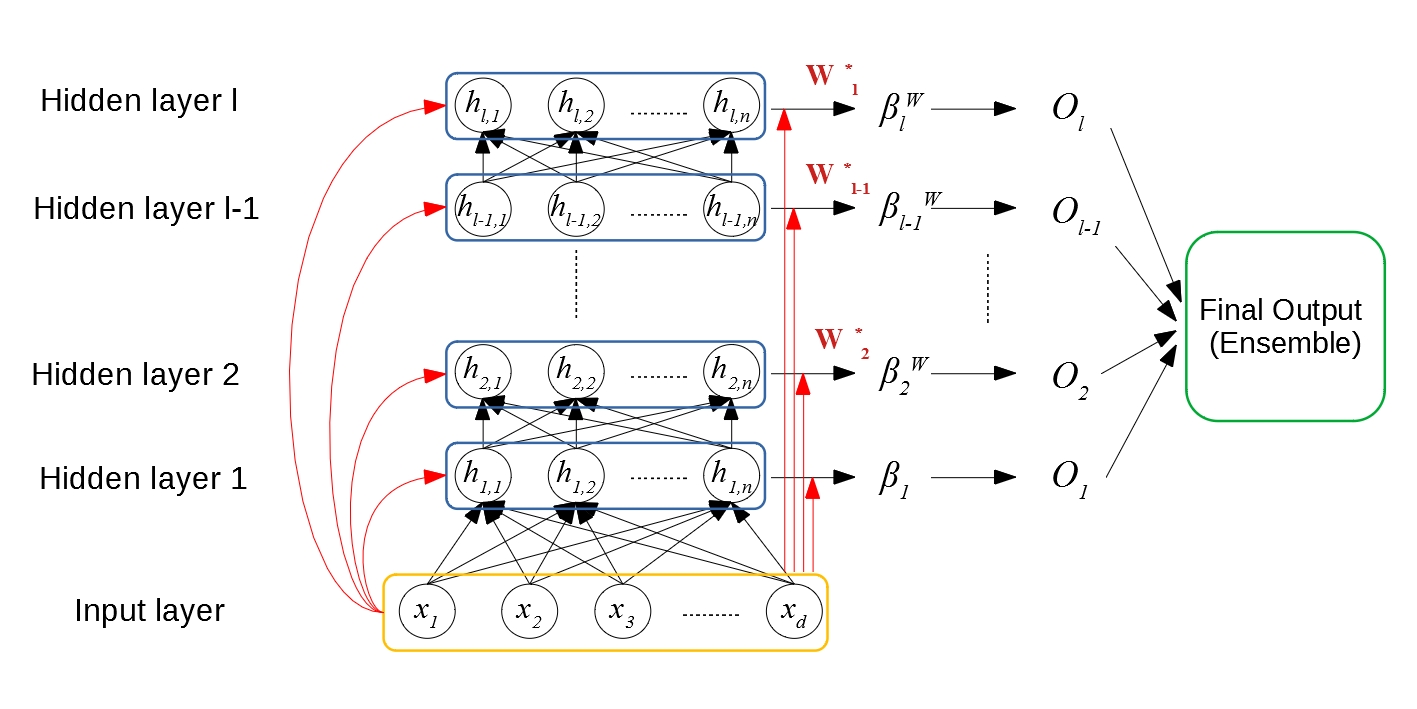}
\caption{The structure of WedRVFL network. From $l\geq 1$, it gives each training sample different importance values when calculating $\bm{\beta}$. This method makes every hidden layer has its preference for a particular range of the samples.}
\label{fig:fig4}
\end{figure}

Weighting \citep{onan2016multiobjective} is a widely used method in boosting ensemble learning and can make each classifier focus its preference on a particular set of samples. Therefore, we propose our own weighting methods in this paper. In our Weighted Ensemble Deep Random Vector Functional Link Network (WedRVFL), we apply weighting by treating each hidden layer as an independent classifier. The samples which are hard to predict will be given higher weights in the next classifier. The typical structure of the WedRVFL network is shown in Fig. \ref{fig:fig4}.
The generation step for the WedRVFL is exactly the same as the basic edRVFL using \eqref{equation:generation1}\eqref{equation:generation2}. Moreover, the way of calculating the output weights $\bm{\beta_1}$ of the first hidden layer is also based on \eqref{equation:primal} and \eqref{equation:dual} (Ridge regression is chosen here). Suppose there are $m$ training samples, after finishing the training for the first layer, some of these samples are predicted correctly while others are assigned wrong labels in this layer's classification. Denote these two groups of samples as $S_r$ and $S_w$, respectively. And let the size of $S_r$ and $S_w$ to be $n_r$ and $n_w$. Then we can have:
\begin{equation}
n_r + n_w = m
\end{equation}
For these samples in set $S_w$, which means they are predicted wrongly in the current layer, we should give them a higher importance value in the next layer. On the other hand, for these samples which have correct classification, their importance value should be decreased in the next classifier. 

In the first hidden layer, there is no weighting scheme for different samples. However, we still can treat this situation as that all the training samples are sharing the same weight 1. From the second hidden layer, we give the samples in set $S_r$ weight $\omega_r$ and samples in set $S_w$ weight $\omega_w$. Since we know that $\omega_r$ should be less than 1 and a positive value. The range of $\omega_r$ is $(0,1]$ ($\omega_r = 1$ if all the samples are predicted correctly in the last layer). Recall that when all the samples are sharing the same weight 1, the sum of these weights is equal to the size of the training set $m$. For numerical stability, we should follow the same rule in our weighting scheme. Therefore, we can obtain the following equation:

\begin{equation}
n_r \times \omega_r + n_w \times \omega_w = m
\end{equation}

So that the weight $\omega_w$ for set $S_w$ can be expressed by $\omega_r$ as:

\begin{equation}
\omega_w = \frac{m - n_r \times \omega_r}{n_w}
\label{equation:weight}
\end{equation}

It is worth mentioning that although we have two different weights for $S_r$ and $S_w$, we only need to set one hyperparameter $\omega_r$ during the experiments. The validation set can be used to find the best setting for this hyperparameter.

After getting the value of $\omega_w$, we can create the weight matrix $\mathbf{W_l}$, where $\mathbf{W_l}\in\mathbb{R}^{m\times 1}$ and $l\geq2$ (weighting scheme starts from the second hidden layer). The $i$th value in $\mathbf{W_l}$ represents the weight we give to the $i$th training sample (either $\omega_r$ or $\omega_w$). Thus, when $l\geq 2$, \eqref{equation:primal} and \eqref{equation:dual} will turn into new forms:
\begin{equation}
\mbox{Primal Space: } \bm{\beta_l^W}=(\mathbf{D^TW_l^*D}+{\lambda}\mathbf{I})^{-1}\mathbf{D^TW_l^*}\mathbf{Y}
\label{equation:W1}
\end{equation}
\begin{equation}
\mbox{Dual Space: } \bm{\beta_l^W}=\mathbf{D^T}(\mathbf{W_l^*DD^T}+{\lambda}\mathbf{I})^{-1}\mathbf{W_l^*Y}
\label{equation:W2}
\end{equation}
where $\mathbf{W_l}^*=
\left[ \begin{array}{cccc}
\mathbf{W_l^{(1)}} & 0 & \cdots & 0\\
0 & \mathbf{W_l^{(2)}} & \cdots & 0\\
\vdots & \vdots & \ddots & \vdots\\
0 & 0 & \cdots & \mathbf{W_l^{(m)}} 
\end{array} 
\right ]$, which is the $(m\times m)$ diagonal form of $\mathbf{W_l}$.
\par
From the above, we know that the training samples which are difficult to train will be given higher importance values when calculating the loss in the next layer. This method ensures that for each training sample, whether it is hard to train or not, there will exist some corresponding layers that are good at predicting such samples. Hence, every sample may have been predicted correctly in some layers. Intuitively, the wrong prediction of WedRVFL can be less than the normal edRVFL after ensemble aggregation. This claim is supported by our empirical simulation results in the experimental part.
\subsection{Pruning based Ensemble Deep Random Vector Functional Link Networks}
\begin{figure}[h!]
\centering
\includegraphics[scale=0.4]{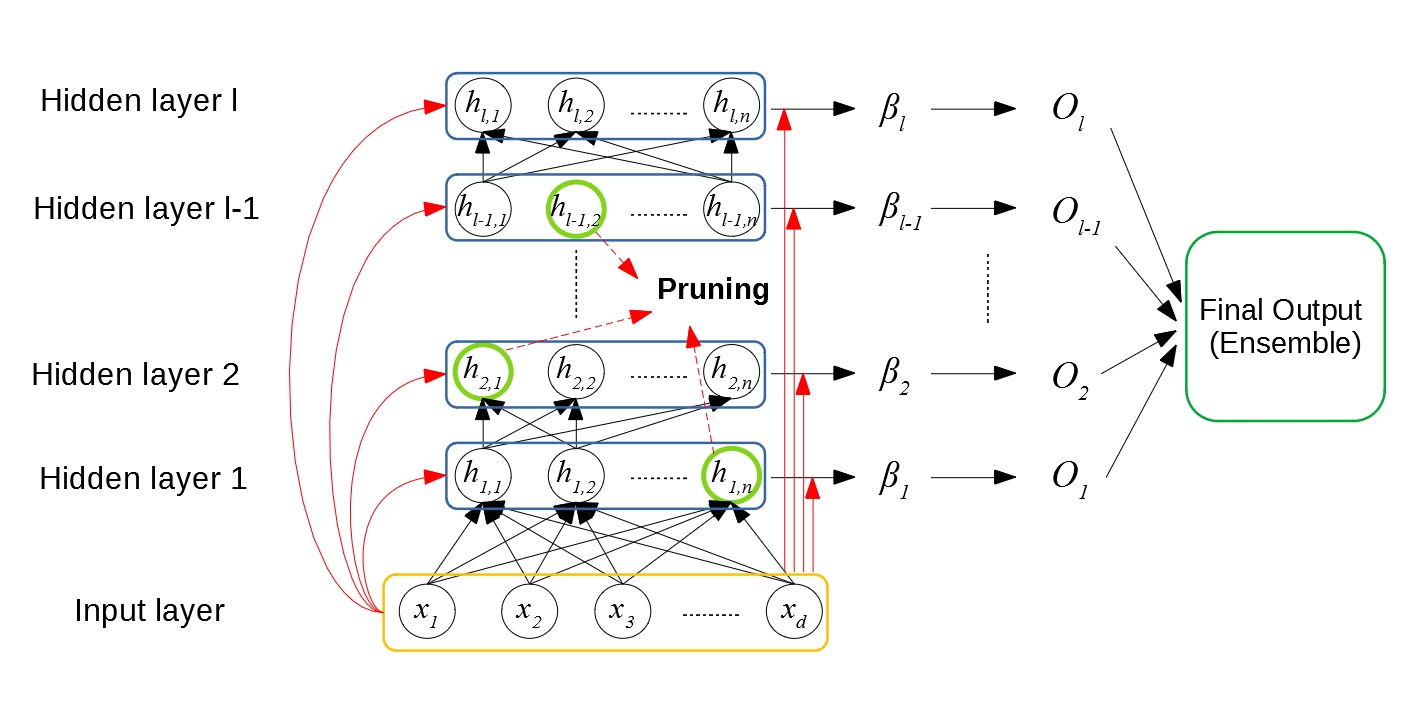}
\caption{The structure of PedRVFL network. It chooses to cut off some inferior neurons starting from the first hidden layer. Here $h_{1,n}$, $h_{2,1}$, and $h_{l-1,2}$ are identified as inferior ones and will be removed.}
\label{fig:fig5}
\end{figure}

The pruning method is widely used in today's neural networks. It was proposed to improve the efficiency of the neural network models when facing a limited computational budget. However, we apply the pruning method in our model for a completely different reason. Since the weights between hidden layers in our neural network are randomly generated and kept fixed, we believe some inferior features will be created and propagated to deeper layers. With the layer number increasing, the accuracy of a single layer's prediction will slightly go down. Hence, we decide to cut off some inferior hidden neurons to prevent them from participating in the generation of the deeper hidden layers. The structure of PedRVFL is shown in Fig.\ref{fig:fig5}.
\par
For deciding which neurons should be cut off, \citep{lecun1989optimal} proposed a pruning method based on the sensitivity calculation. Besides, Penalty-term methods are also widely used by researchers \citep{chauvin1988back,ji1990generalizing}. In recent days, some new pruning schemes targeted on deep neural networks have been investigated \citep{han2015learning,molchanov2016pruning}. In this paper, we set a criterion based on the output weights which also belongs to the sensitivity methods. For each hidden neuron $h_{l,n}$ in layer $l$, it has $k$ corresponding weights in the $\bm{\beta_l}$ where $k$ refers to the number of classes. Let these weights be $a_{l,n,1},a_{l,n,2}\dots a_{l,n,k}$, and the absolute value for them represent the importance for this neuron to different classes. Therefore, we should use the sum of these values to evaluate how much is this hidden neuron contributed to the whole prediction:
\begin{equation}
\theta_{l,n}=\sum_{i=1}^{k}|a_{l,n,i}|
\label{equation:Pruning}
\end{equation}
Depending on the pruning rate we set, hidden neurons with the lowest values of $\theta$ are pruned. It is worth mentioning that by applying the pruning method, our neural network will have fewer hidden neurons. That means the complexity of the framework is also reduced. However, according to the experimental results, this model turns out to have better performance on multiple tasks.
\subsection{Weighting and Pruning based Ensemble Deep Random Vector Functional Link Network}
\begin{figure}[h!]
\centering
\includegraphics[scale=0.4]{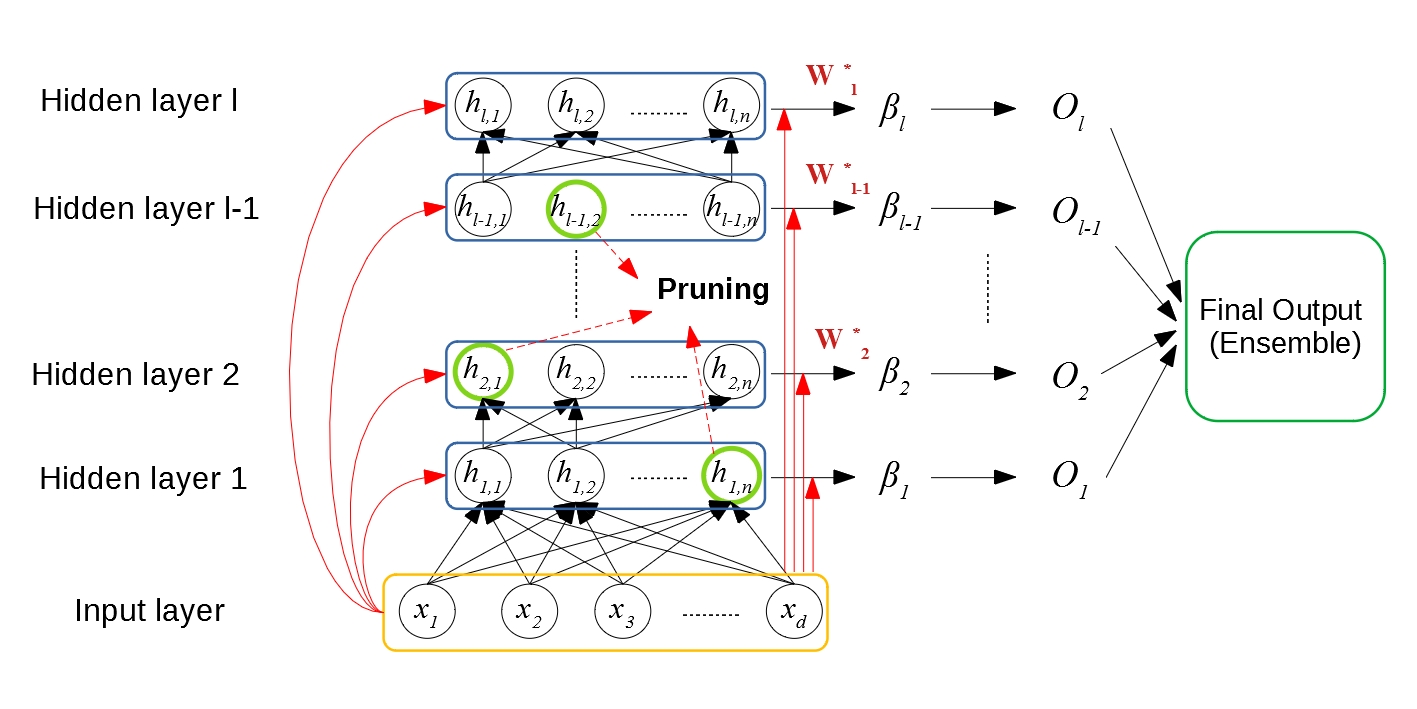}
\caption{The structure of WPedRVFL network. It combines the weighting method and the pruning method to achieve better performance.}
\label{fig:fig6}
\end{figure}

Weighting and Pruning based Ensemble Deep Random Vector Functional Link Network (WPedRVFL) is a combination of the above two models. It has both advantages of WedRVFL and PedRVFL. The structure of this neural network is shown in Fig.\ref{fig:fig6}.

To better illustrate the training procedure of the WPedRVFL, we summarize the steps in Algorithm \ref{al:1}.

It is also worth highlighting that a BP-trained network needs to re-train all the parameters in the previously trained hidden layers every time if we set different number of the hidden layers. However, in our edRVFL based neural networks, we only need to train the model once with the maximum permitted hidden layer number. Since there is no need for training the hidden weights and biases in edRVFL based neural networks, results of different numbers of hidden layers can be obtained by eliminating the last few layers' predictions in the final output.

\begin{algorithm}[h!]
\DontPrintSemicolon
\SetAlgoLined
\SetKwInOut{Input}{Input}
\SetKwRepeat{Do}{do}{while}
\Input{A set of training samples $\mathbf{X} \in \mathbb{R}^{m\times d}$.}
 \For{Every Hidden Layer $l$}{
    Initialize the hidden weights.\;
    Do re-normalization and generate the hidden features using \eqref{equation:re}, \eqref{equation:generation1} and \eqref{equation:generation2}.\;
    \textbf{Weighting:} Get the optimal output weights $\bm{\beta_l^W}$ by \eqref{equation:W1} and \eqref{equation:W2}. When $l=1$, all the samples share the same weight 1.\;
    Predict the labels of the training samples.\;
    \textbf{Weighting:} Calculate the weight $\omega_w$ for the wrong predicted samples based on the $\omega_r$ we set by \eqref{equation:weight}. \;
    \textbf{Pruning:} Obtain the importance value $\theta$ for each hidden nodes using \eqref{equation:Pruning}.\;
    \textbf{Pruning:} Cut off a number of the inferior neurons according to the pruning rate $p$ we set.
 }
 Combine all the predictions from every hidden layer using ensemble methods.\;
 \SetKwInOut{Output}{Output}
 \Output{
 Hidden weights, the final prediction of the training samples, and the output weights $\bm{\beta_l^W}$ of every hidden layer.
 }
 \caption{WPedRVFL}
 \label{al:1}
\end{algorithm}
\section{Experiments}
\subsection{Datasets}
In this paper, we use 24 classification datasets from the UCI machine learning repository \citep{asuncion2007uci} to compare the performance of our methods with other state-of-the-art networks. These datasets are from diverse application areas like physics, geology, and biology and usually serve as the benchmarks for the classification task \citep{asuncion2007uci,klambauer2017self}. We follow the same data pre-processing and partitions as used in \citep{klambauer2017self}. The details of these datasets are shown in Table \ref{table:table1}. 
\newpage
\newcommand{\myfnmark}[1]{\mbox{\textsuperscript{\normalfont #1}}}
\makeatletter
\renewcommand\@makefntext[1]%
   {\noindent\makebox[1.8em][r]{\myfnmark}#1} 
\makeatother
    \begin{longtable}{lrrr}
    \caption{Datasets used in this work}
    \label{table:table1}
    \\
    Dataset & \multicolumn{1}{l}{\#Patterns} & \multicolumn{1}{l}{\#Features} & \multicolumn{1}{l}{\#Classes} \\\hline
    \endfirsthead
    
    \multicolumn{3}{l}{\textbf{\tablename \ \thetable{}}.  \textsl{(Continued)}} \\
    \\
    Dataset & \multicolumn{1}{l}{\#Patterns} & \multicolumn{1}{l}{\#Features} & \multicolumn{1}{l}{\#Classes} \\\hline
    \endhead
    
    \hline
    \multicolumn{4}{r}{\textsl{(Continued)}}\\
    \endfoot
    
    \hline
    \endlastfoot
    abalone & 4177  & 9     & 3 \\
    adult & 48842 & 15 & 2 \\
    arrhythmia & 452   & 263   & 13 \\
    bank  & 4521  & 17    & 2 \\
    cardio-10 & 2126  & 22    & 10 \\
    cardio-3 & 2126  & 22    & 3 \\
    chess-krvkp & 3196  & 37    & 2 \\
    congressional-voting & 435   & 17    & 2 \\
    contrac & 1473  & 10    & 3 \\
    glass & 214   & 10    & 6 \\
    letter & 20000 & 17    & 26 \\
    molec-biol-splice & 3190  & 61    & 3 \\
    monks-3 & 554   & 7     & 2 \\
    musk-2 & 6598  & 167   & 2 \\
    oocytes\_trisopterus\_states\_5b & 912   & 26    & 2 \\
    pendigits & 10992 & 17 & 10 \\
    spambase & 4601  & 58    & 2 \\
    statlog-image & 2310  & 19    & 7 \\
    statlog-landsat & 6435 & 37 & 6 \\
    statlog-shuttle & 58000 & 10 & 7 \\
    wall-following & 5456  & 25    & 4 \\
    waveform & 5000  & 22    & 3 \\
    waveform-noise & 5000  & 41    & 3 \\
    wine-quality-white & 4898  & 12    & 7 \\
    \hline
    \end{longtable}
\begin{minipage}{.8\linewidth}
\begin{tablenotes}
    \item \myfnmark{$\ast$} We name these datasets the same as in \citep{klambauer2017self}.
\end{tablenotes}
\end{minipage}
\newpage
\subsection{Other methods used for comparison}

For evaluating the performance of our proposed edRVFL variants, we compare them with other 4 deep learning models, 3 RVFL based randomized neural networks, and the original edRVFL network without re-normalization in this work. These methods (1-8), as well as our new proposed methods (9-12), are listed as follows:
\begin{enumerate}[1)]
\item Self Normalizing Neural Network (SNN): Self normalizing networks with SELUs activation function and ranks the best among the FNNs \citep{klambauer2017self}.
\item MSRAinit (MS): The deep neural network with Microsoft weight initialization \citep{He_2015_ICCV}.
\item Highway (HW): Highway networks \citep{NIPS2015_5850}.
\item ResNet (ResNet): Residual networks adapted to FNNs using residual blocks with 2 or 3 layers \citep{he2016deep}.
\item Stochastic configuration networks (SCN): The
randomized neural network generated incrementally by stochastic configuration\citep{wang2017stochastic}.
\item Broad learning system (BLS): The RVFL based model with a complete paradigm shift in discriminative learning \citep{chen2017broad}.
\item Hierarchical ELM (H-ELM): Hierarchical Extreme Learning Machine for multi-layer perception \citep{tang2015extreme}
\item edRVFL\_O: The original ensemble deep random vector functional link neural network without re-normalization \citep{Shi2021Random}.
\item edRVFL\_N: The new edRVFL network with re-normalization that proposed in this work.
\item PedRVFL: Pruning based edRVFL with re-normalization proposed in this work.
\item WedRVFL: Weighting based edRVFL with re-normalization proposed in this work.
\item WPedRVFL: The combination of PedRVFL and WedRVFL with re-normalization proposed in this work.
\end{enumerate}

\subsection{Hyperparameter Settings}
\indent
For these FNN methods, we use the same hyperparameter settings as in \citep{klambauer2017self}. For the other RVFL based randomized neural networks, We use the official codes provided by their authors and set the ranges of the hyperparameters as suggested in their original papers. Moreover, for the edRVFL based methods, the regularization parameter $\lambda$ is chosen from range $2^x$, where $x$ belongs to $[-12,12]$. The hidden neuron number $n$ of these methods can be tuned within $[20,1000]$. The maximum number of the hidden layers $l_{max}$ of the edRVFL based methods is set to 10. The two key hyperparameters $\gamma$ and $\alpha$ of the re-normalization are tuned among $[0.5,2]$ and $[-2,2]$, respectively. The WedRVFL can choose the best weight $\omega_r$ in $(0,1]$ based on the validation accuracy. Also, for pruning-based methods, the pruning rate $p$ is tuned among $[0,1)$. The details about these settings are summarized in Table \ref{table:parameters}. 
\par

During the tuning process, we do 4-fold cross-validation to find the best parameter settings. We separate the whole dataset into training and testing sets 4 times. In each fold, 25$\%$ of the training data are used as the validation set, and we select the hyperparameter configuration with the best average validation accuracy. Then, we use the whole training data to re-train the models before feeding the test data into them. The testing accuracy is obtained based on the correct predictions of the networks for the test data. At last, we report the mean value of the 4 testing accuracy as the final classification result for the current dataset.
\par
In order to test the robustness of our new methods, for each edRVFL based network, we run the above experiment 10 times with different randomized hidden features. Then, we report the mean value and the standard deviation of these 10 outcomes in Table \ref{table:results}.

\begin{table}[h!]
    \begin{threeparttable}
    \caption{Hyperparameters considered for edRVFL based methods} 
    \label{table:parameters}
    \centering
    \begin{tabular}{l p{0.5\columnwidth}}\toprule
        Hyperparameter & Considered values    \\ \midrule
        Regularization parameter $\lambda$ & $\lambda$ belongs to $2^{x}$, $x \in [-12,12]$ \\ Number of hidden neurons $n$ & [20,1000]\\
        Maximum number of hidden layers $l_{max}$ & 10  \\
        $\gamma$ in batch normalization & [0.5,2] \\
        $\alpha$ in batch normalization & [-2,2] \\
        Weight $\omega_r$ for the correctly predicted samples & $(0,1]$, 1 means there is no weighting in the network\\
        Pruning Rate $p$ & [0,1), 0 means there is no pruning in the network \\
        \bottomrule
        \end{tabular}
    \end{threeparttable}    
\end{table}

\subsection{Experimental Results}
The performance of all the 12 methods on 24 UCI datasets is shown in Table \ref{table:results}.
\newgeometry{left=2cm,right=1cm,top=2cm,bottom=2cm}
{
\begin{landscape}
{
    \footnotesize
    \begin{longtable}{lccccccccccccc}
    \caption{Comparison of Accuracy(\%) on 24 UCI Datasets}
    \label{table:results}
    \\
    Dataset & \multicolumn{1}{l}{SNN\citep{klambauer2017self}} & \multicolumn{1}{l}{MS\citep{He_2015_ICCV}} & \multicolumn{1}{l}{HW\citep{NIPS2015_5850}} & \multicolumn{1}{l}{Resnet\citep{he2016deep}} &
    \multicolumn{1}{l}{SCN\citep{wang2017stochastic}} &
    \multicolumn{1}{l}{BLS\citep{chen2017broad}} &
    \multicolumn{1}{l}{H-ELM\citep{tang2015extreme}} & 
    \multicolumn{1}{l}{edRVFL\_O\citep{Shi2021Random}} &
    \multicolumn{1}{l}{edRVFL\_N$^{\dagger}$} &
    \multicolumn{1}{l}{WedRVFL$^{\dagger}$} &
    \multicolumn{1}{l}{PedRVFL$^{\dagger}$} &
    \multicolumn{1}{l}{WPedRVFL$^{\dagger}$} & \\\hline
    \endfirsthead
    
    \multicolumn{3}{l}{\textbf{\tablename \ \thetable{}}.  \textsl{(Continued)}} \\
    \\
    Dataset & \multicolumn{1}{l}{SNN\citep{klambauer2017self}} & \multicolumn{1}{l}{MS\citep{He_2015_ICCV}} & \multicolumn{1}{l}{HW\citep{NIPS2015_5850}} & \multicolumn{1}{l}{Resnet\citep{he2016deep}} &
    \multicolumn{1}{l}{SCN\citep{wang2017stochastic}} &
    \multicolumn{1}{l}{BLS\citep{chen2017broad}} &
    \multicolumn{1}{l}{H-ELM\citep{tang2015extreme}} & 
    \multicolumn{1}{l}{edRVFL\_O\citep{Shi2021Random}} &
    \multicolumn{1}{l}{edRVFL\_N$^{\dagger}$} &
    \multicolumn{1}{l}{WedRVFL$^{\dagger}$} &
    \multicolumn{1}{l}{PedRVFL$^{\dagger}$} &
    \multicolumn{1}{l}{WPedRVFL$^{\dagger}$} & \\\hline
    \endhead
    
    \hline
    \multicolumn{4}{r}{\textsl{(Continued)}}\\
    \endfoot

    \hline
    \endlastfoot
    abalone & 66.57  & 62.84  & 64.27  & 64.66  & 64.00  & 60.06  & 63.77  & 65.83±0.36  & 66.13±0.32  & 66.87±0.22  & \textbf{67.05±0.01}  & 66.89±0.15  \\
    adult & 84.76  & 84.87  & 84.53  & 84.84  & 85.01  & 85.15  & 85.05  & 85.21±0.15  & 85.26±0.15  & 85.30±0.13  & 85.42±0.14  & \textbf{85.46±0.14}  \\
    arrhythmia & 65.49  & 63.72  & 62.83  & 64.60  & 44.91  & 62.23  & 72.12  & 69.06±0.69  & 72.43±0.66  & 73.22±0.32  & 73.66±0.39  & \textbf{73.88±0.36}  \\
    bank  & 89.03  & 88.76  & 88.85  & 87.96  & 88.83  & 88.19  & 89.20  & 89.77±0.11  & 90.13±0.11  & 90.92±0.11  & 90.19±0.11  & \textbf{91.14±0.01}  \\
    cardio-10 & 83.99  & 84.18  & 84.56  & 81.73  & 81.26  & 83.47  & 82.39  & 82.37±0.47  & 83.24±0.42  & 84.56±0.52  & 84.08±0.26  & \textbf{85.30±0.26}  \\
    cardio-3 & 91.53  & 89.64  & 91.71  & 90.21  & 91.57  & 91.33  & 90.68  & 92.71±0.25  & 93.42±0.25  & 93.47±0.21  & 93.55±0.01  & \textbf{94.20±0.18}  \\
    chess-krvkp & 98.37  & 99.00  & 99.00  & 99.12  & 97.77  & 98.75  & 99.00  & 99.08±0.15  & 99.21±0.17  & 99.36±0.01  & 99.33±0.01  & \textbf{99.47±0.01}  \\
    congressional-voting & 61.47  & 60.55  & 58.72  & 59.63  & 60.09  & 59.40  & 61.24  & 61.01±0.28  & 61.98±0.29  & 61.91±0.23  & 61.93±0.27  & \textbf{62.16±0.25}  \\
    contrac & 51.90  & 51.36  & 50.54  & 51.36  & 47.75  & 41.78  & 54.08  & 51.33±0.50  & 54.04±0.49  & 55.53±0.51  & 54.94±0.37  & \textbf{55.68±0.19}  \\
    glass & \textbf{73.58}  & 60.38  & 64.15  & 64.15  & 66.79  & 61.23  & 68.87  & 65.13±0.88  & 70.73±0.83  & 71.75±0.67  & 71.19±0.47  & 72.35±0.87  \\
    letter & 97.26  & 97.12  & 89.84  & 97.62  & 86.11  & 93.99  & 93.15  & 97.43±0.23  & 97.54±0.22  & \textbf{97.77±0.15}  & 97.66±0.07  & 97.73±0.11  \\
    molec-biol-splice & 83.72  & 84.82  & \textbf{88.33}  & 85.57  & 75.75  & 74.84  & 82.40  & 84.01±0.38  & 84.31±0.39  & 84.07±0.50  & 85.22±0.22  & 85.69±0.57  \\
    monks-3 & 60.42  & 74.54  & 58.80  & 58.33  & 69.42  & 52.37  & 78.70  & 55.02±2.02  & 75.48±2.04  & 80.29±1.93  & 81.98±1.12  & \textbf{82.35±1.57}  \\
    musk-2 & 98.03  & 99.45  & 99.15  & 99.64  & 96.71  & 98.77  & 98.32  & 98.54±0.25  & 99.33±0.26  & 99.54±0.19  & 99.28±0.01  & \textbf{99.74±0.01}  \\
    oocytes\_trisopterus\_states\_5b& 93.42  & 94.30  & 93.42  & 89.47  & 89.91  & 57.46  & 92.06  & 93.91±0.20  & 93.97±0.20  & 94.19±0.18  & 95.20±0.23  & \textbf{95.21±0.30}  \\
    pendigits & 97.06  & 97.14  & 96.71  & 97.08  & 97.05  & 97.45  & 97.41  & 97.49±0.12  & 97.97±0.13  & 98.05±0.15  & 97.89±0.11  & \textbf{98.20±0.11}  \\
    spambase & 93.00  & 94.61  & 94.35  & 94.61  & 91.71  & 92.15  & 92.67  & 93.83±0.16  & 94.08±0.17  & 94.11±0.23  & 94.18±0.01  & \textbf{94.72±0.14}  \\
    statlog\_image & 95.49  & \textbf{97.57}  & 95.84  & 95.84  & 94.97  & 89.90  & 95.28  & 96.82±0.21  & 97.05±0.17  & 97.56±0.14  & 97.44±0.18  & 97.40±0.13  \\
    statlog\_landsat & 91.00  & 90.75  & 91.10  & 90.55  & 90.25  & 83.47  & 91.22  & 91.15±0.51  & 91.64±0.51  & \textbf{92.19±0.25}  & 92.15±0.17  & 91.85±0.12  \\
    statlog\_shuttle & 99.90  & 99.83  & 99.77  & 99.92  & 99.79  & 96.82  & 99.88  & 99.91±0.02  & 99.92±0.02  & 99.93±0.01  & 99.93±0.01  & \textbf{99.94±0.01}  \\
    wall-following & 90.98  & 90.76  & \textbf{92.30}  & 90.12  & 85.41  & 89.53  & 89.46  & 90.28±0.43  & 90.79±0.45  & 91.04±0.55  & 91.37±0.31  & 92.20±0.33  \\
    waveform & 84.80  & 83.12  & 83.20  & 83.60  & 84.76  & 83.48  & 86.16  & 85.95±0.11  & 85.97±0.11  & 86.59±0.01  & 86.83±0.01  & \textbf{86.97±0.12}  \\
    waveform-noise & 86.08  & 83.28  & 86.96  & 85.84  & 83.70  & 82.44  & 86.08  & 85.68±0.14  & 86.17±0.11  & 86.92±0.01  & 86.98±0.15  & \textbf{87.13±0.13}  \\
    wine-quality-white & 63.73  & 64.79  & 55.64  & 63.07  & 55.96  & 55.15  & 55.49  & 63.29±0.40  & 63.70±0.39  & 64.76±0.38  & 64.22±0.41  & \textbf{65.66±0.31}  \\
    \hline
    \textbf{Mean Accuracy} & 83.40  & 83.22  & 82.27  & 82.48  & 80.40  & 78.31  & 83.53  & 83.12±0.38  & 84.77±0.37  & 85.41±0.32  & 85.49±0.21  & \textbf{85.89±0.27}  \\
    \textbf{Ave. Rank} & 7.29  & 7.42  & 7.83  & 7.85  & 10.08  & 10.42  & 7.77  & 7.00  & 4.85  & 3.04  & 3.02  & \textbf{1.42}  \\
    \hline
    \end{longtable}
\begin{minipage}{.8\linewidth}
\begin{tablenotes}
     \item \myfnmark{$\ast$} Methods with $^{\dagger}$ are proposed in this paper with re-normalization.
\end{tablenotes}
\end{minipage}}
\end{landscape}
}
\restoregeometry
\newpage
\begin{table}[h!]
\centering
\begin{threeparttable}[b]
\caption{Statistical comparison between WPedRVFL and each of the other networks}
\label{table:table4}
\begin{tabular}{lll}
\hline
Methods & Ave. Rank & $p$-value \\
\hline
WPedRVFL & 1.42 &         \\
PedRVFL  & 3.02     & 1.7e-1   \\ 
WedRVFL$^*$ & 3.04      & 6.7e-4    \\
edRVFL\_N$^*$  & 4.85      & 2.5e-3    \\
edRVFL\_O$^*$ \citep{Shi2021Random} & 7.00   & 1.7e-3    \\
SNN$^*$ \citep{klambauer2017self} & 7.29      & 6.7e-4    \\
MS$^*$ \citep{He_2015_ICCV} & 7.42 & 1.1e-4 \\
H-ELM$^*$ \citep{tang2015extreme}                               & 7.77      & 5.6e-5    \\
HW$^*$ \citep{NIPS2015_5850} & 7.83 & 2.4e-5 \\ 
Resnet$^*$ \citep{he2016deep}   & 7.85      & 1.8e-5    \\
SCN$^*$ \citep{wang2017stochastic} & 10.08 & 1.8e-5    \\
BLS$^*$ \citep{chen2017broad} & 10.42 & 1.8e-5    \\
\hline
\end{tabular}
\begin{tablenotes}
     \item[*] Lower rank reﬂects better performance. The $p-$ value is obtained from the paired Wilcoxon test. Methods that are signiﬁcantly worse than the best method are marked with “*”.
\end{tablenotes}
\end{threeparttable}
\end{table}
\begin{table}[h!]
\centering
\begin{threeparttable}
\setlength{\abovecaptionskip}{0pt}%
\setlength{\belowcaptionskip}{10pt}%
\caption{Statistical Comparison of all the Methods}
\label{table:wilcoxon_all}
\begin{tabular}{|l|l|l|l|l|l|l|l|l|l|l|l|l|}
\hline
& \rotatebox{90}{WPedRVFL$^{\dagger}$} & 
\rotatebox{90}{PedRVFL$^{\dagger}$} & 
\rotatebox{90}{WedRVFL$^{\dagger}$} &
\rotatebox{90}{edRVFL\_N$^{\dagger}$} &
\rotatebox{90}{edRVFL\_O\citep{Shi2021Random}} &
\rotatebox{90}{SNN\citep{klambauer2017self}} &
\rotatebox{90}{MS\citep{He_2015_ICCV}} &
\rotatebox{90}{H-ELM\citep{tang2015extreme}} & 
\rotatebox{90}{HW\citep{NIPS2015_5850}} & 
\rotatebox{90}{Resnet\citep{he2016deep}} &
\rotatebox{90}{SCN\citep{wang2017stochastic}} &
\rotatebox{90}{BLS\citep{chen2017broad}}\\ \hline
WPedRVFL$^{\dagger}$ &   &   & + & + & + & + & + & + & + & + & + & + \\ \hline
PedRVFL$^{\dagger}$ &   &   &   & + & + & + & + & + & + & + & + & + \\ \hline
WedRVFL$^{\dagger}$ & - &   &   &   &   &   & + & + & + & + & + & + \\ \hline
edRVFL\_N$^{\dagger}$ & - & - &   &   &   &   &   &   & + & + & + & + \\ \hline
edRVFL\_O\citep{Shi2021Random} & - & - &   &   &   &   &   &   & + & + & + & + \\ \hline
SNN\citep{klambauer2017self} & - & - &   &   &   &   &   &   & + & + & + & + \\ \hline
MS\citep{He_2015_ICCV} & - & - & - &   &   &   &   &   & + & + & + & + \\ \hline
H-ELM\citep{tang2015extreme} & - & - & - &   &   &   &   &   & + & + & + & + \\ \hline
HW\citep{NIPS2015_5850} & - & - & - & - & - & - & - & - &   & + & + & + \\ \hline
Resnet\citep{he2016deep} & - & - & - & - & - & - & - & - & - &   &   & + \\ \hline
SCN\citep{wang2017stochastic} & - & - & - & - & - & - & - & - & - &   &   & + \\ \hline
BLS\citep{chen2017broad} & - & - & - & - & - & - & - & - & - & - & - &  \\ \hline
\end{tabular}
\begin{tablenotes}
\item[*] The empty cell in the table means the corresponding method in the row and column have no significant statistical difference. The symbol '+' indicates that the method in the row is statistically better than the method in the column. On the other hand, the symbol '-' means the method in the row is statistically worse than that in the column.
\item[*] Methods with $^{\dagger}$ are proposed in this paper.
\end{tablenotes}
\end{threeparttable}
\end{table}
\par
We take the results of SNN, MS, HW, and Resnet directly from \citep{klambauer2017self}. The rankings of each classifier are used to compare their performance on all the datasets. For the ranking method, the best classifier based on the classification accuracy of one dataset is ranked 1, the second is ranked 2, and so on. Then we use the average rank of all 24 datasets to show the performance of the classifiers. The best results for each dataset, as well as the best global rank and accuracy, are given in bold.
\par
Wilcoxon signed-rank test is a non-parametric statistical hypothesis test used to compare the performance of two related measurements on multiple tasks \citep{mann1947test}. In this paper, we employ it to do the pairwise comparison on the selected two methods to investigate the statistical difference between them. We first compare all the other 11 methods' performance with WPedRVFL. And the results are shown in Table \ref{table:table4}. From the table, we can see that all leading positions are occupied by edRVFL based methods. The top-ranked method is WPedRVFL and followed by WedRVFL. PedRVFL takes the third place, and edRVFL\_N, edRVFL\_O are following behind. By performing the statistical comparison, we know that except for PedRVFL network, WPedRVFL significantly outperforms all the other competitors. 
\par
We also use the Wilcoxon test to do the pairwise comparison between all the 12 classifiers in this paper. The results are shown in Table \ref{table:wilcoxon_all}. If there is no statistically significant difference between the methods in the corresponding row and column, the cells where they intersect will be empty. If the symbol in the cell is ‘+’, that means the method in the corresponding row is statistically better than that in the column. On the contrary, the symbol ‘-’ indicates that the method in the corresponding row is statistically worse.
\par
\begin{table}[h!]
\centering
\begin{threeparttable}[b]
\caption{Standard deviation comparison between edRVFL based methods}
\label{table:std}
\begin{tabular}{ll}
\hline
Methods & Ave. std \\
\hline
PedRVFL  & 0.21\\ 
WPedRVFL & 0.27        \\

WedRVFL$^*$ & 0.32    \\
edRVFL\_N$^*$  & 0.37    \\
edRVFL\_O$^*$ \citep{Shi2021Random} & 0.38\\
\hline
\end{tabular}
\end{threeparttable}
\end{table}
Then, we pay attention to the standard deviation of these edRVFL based methods. We summarize their performance in Table \ref{table:std}. From it, we can learn that our new methods WedRVFL, PedRVFL, and WPedRVFL are more robust than the basic edRVFL\_O and edRVFL\_N. Among them, PedRVFL has the smallest std value. Since WPedRVFL is the combination of WedRVFL and PedRVFL, so we believe this is the reason why its std value is between WedRVFL and PedRVFL.
\par
From all the Tables above, we know that our new proposed methods, edRVFL\_N, WedRVFL, PedRVFL, and WPedRVFL, show great performance on 24 UCI classification tasks. Among them, WPedRVFL takes great advantage both in the average rank and mean accuracy compared to all other methods. It significantly outperforms other FNNs like Resnet, HW, MS, and SNN. And it also has a significant statistical difference from other randomized neural networks. Moreover, the standard deviation of our new methods are smaller than the original edRVFL networks. Therefore, WPedRVFL can be considered as a highly competitive classifier on tabular classification tasks.
\par
\subsection{Effects of the key hyperparameters: Weight $\omega_r$ and Pruning rate $p$}

In this part, we conduct more experiments to discover the effects of the key hyperparmeters. Here we select 4 datasets: arrhythmia, congressional\_voting, statlog\_image, and waveform-noise from the previous section to show how the testing accuracy changes with different settings of $\omega_r$ and $p$.

We use the WPedRVFL model to perform these tests. During the experiments, except for $\omega_r$ and $p$, all the other hyperparameters are fixed to control variables. Moreover, when we are testing the influence of weight $\omega_r$, the pruning rate $p$ is also set to 0 (no pruning) to complete the controlled experiment. On the other hand, the weight $\omega_r$ is set to 1 (no weighting) when the pruning rate $p$ is investigated.  We give the details of these configurations in Table \ref{table:key parameters}. 

\begin{table}[h!]
    \begin{threeparttable}
    \caption{Hyperparameters considered in the controlled experiments} 
    \label{table:key parameters}
    \centering
    \begin{tabular}{l p{0.5\columnwidth}}\toprule
        Hyperparameter & Considered values    \\ \midrule
        Regularization parameter $\lambda$ & 1 \\ Number of hidden neurons $n$ & 500\\
        Number of hidden layers $l$ & 3  \\
        $\gamma$ in batch normalization & 1 \\
        $\alpha$ in batch normalization & 0 \\
        Weight $\omega_r$ for the correctly predicted samples & ${0.2,0.4,0.6,0.8,1}$ or 1 (when testing different pruning rates) \\
        Pruning Rate $p$ & ${0, 0.2,0.4,0.6,0.8}$, or 0 (when testing different weights)\\
        \bottomrule
        \end{tabular}
    \end{threeparttable}    
\end{table}

\begin{figure}[htbp]
\centering
\subfigure[arrhythmia]{
\includegraphics[width=5.5cm]{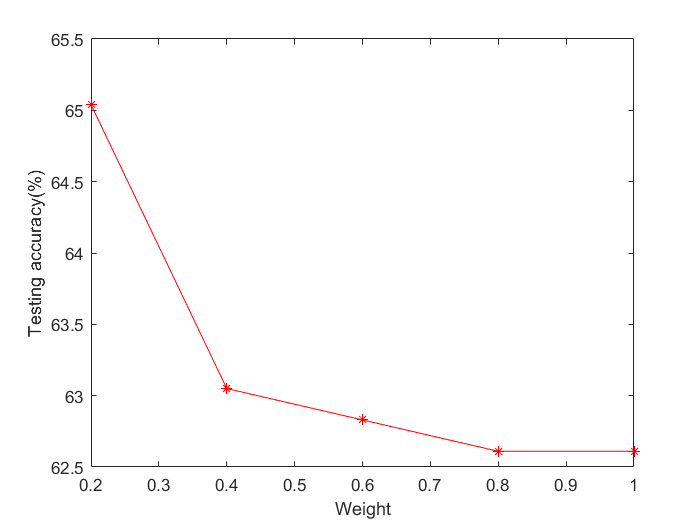}
}
\quad
\subfigure[congressional\_voting]{
\includegraphics[width=5.5cm]{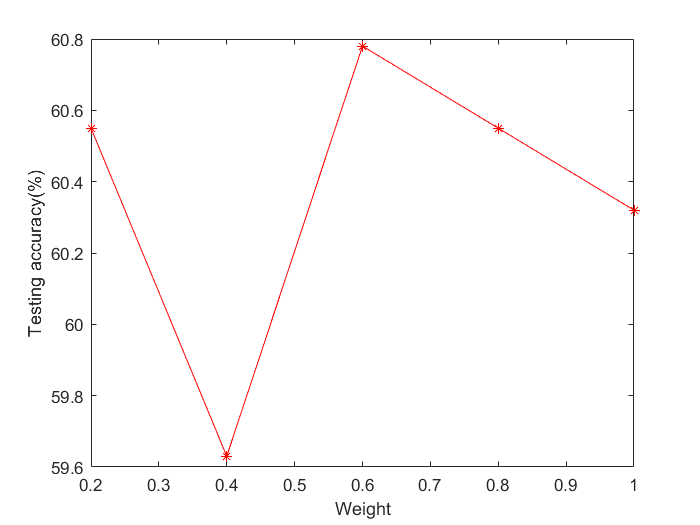}
}
\quad
\subfigure[statlog\_image]{
\includegraphics[width=5.5cm]{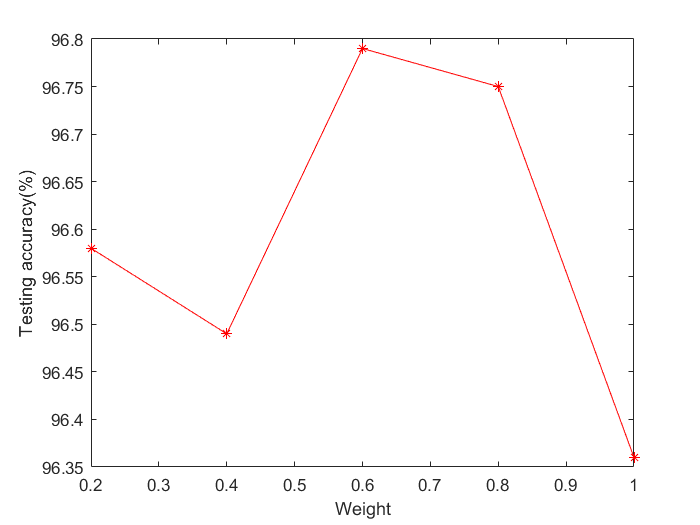}
}
\quad
\subfigure[waveform-noise]{
\includegraphics[width=5.5cm]{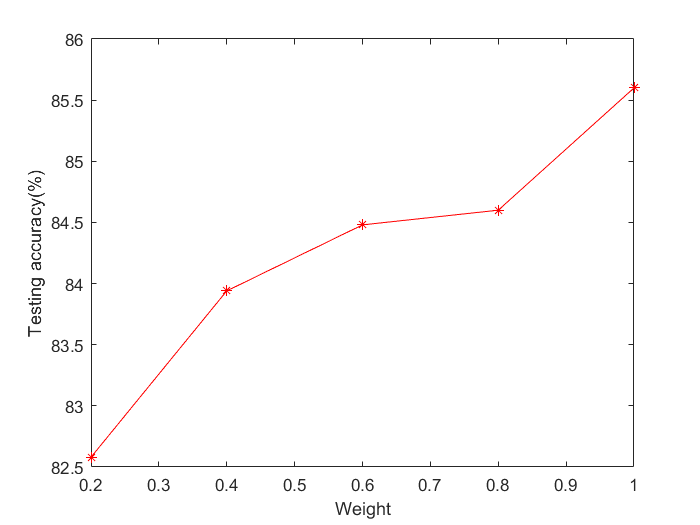}
}
\caption{The testing accuracy of WPedRVFL network with different weight $\omega_w$ on 4 tabular datasets.}
\label{figure:weight}
\end{figure}

We first give WPedRVFL's performance with different weights $\omega_r$ on 4 datasets in Fig. \ref{figure:weight}. As we can see from Fig. \ref{figure:weight}. The weighting scheme is useful in most cases. For arrhythmia, congressional\_voting, and statlog\_image datasets, setting the weight $\omega_r$ for the wrongly predicted samples generally increases the testing accuracy of the model. But which weight $\omega_r$ is the most suitable one varies from case to case. Therefore, we need to use the validation data to find the best configuration of the specific dataset. On the other hand, there is an exception that the weighting scheme is harmful to the classification. For the waveform-noise dataset, the testing accuracy keeps decreasing when we set lower and lower values for $\omega_r$. 

\begin{figure}[htbp]
\centering
\subfigure[arrhythmia]{
\includegraphics[width=5.5cm]{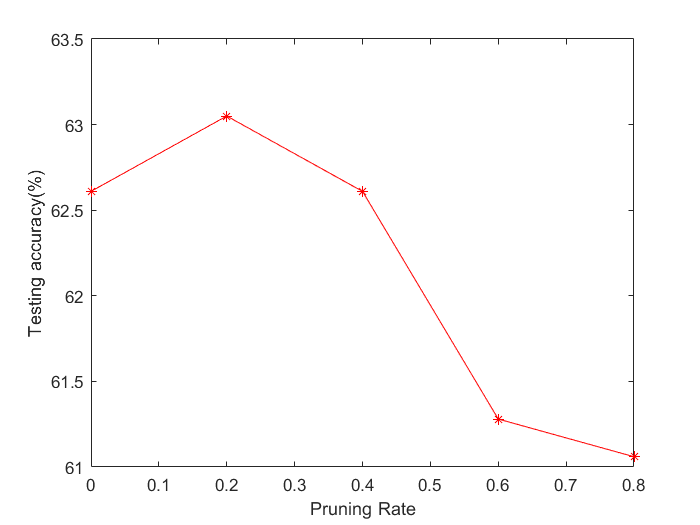}
}
\quad
\subfigure[congressional\_voting]{
\includegraphics[width=5.5cm]{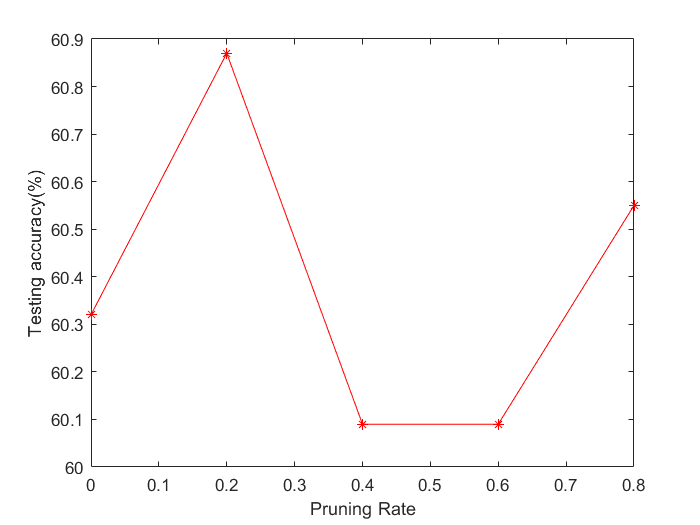}
}
\quad
\subfigure[statlog\_image]{
\includegraphics[width=5.5cm]{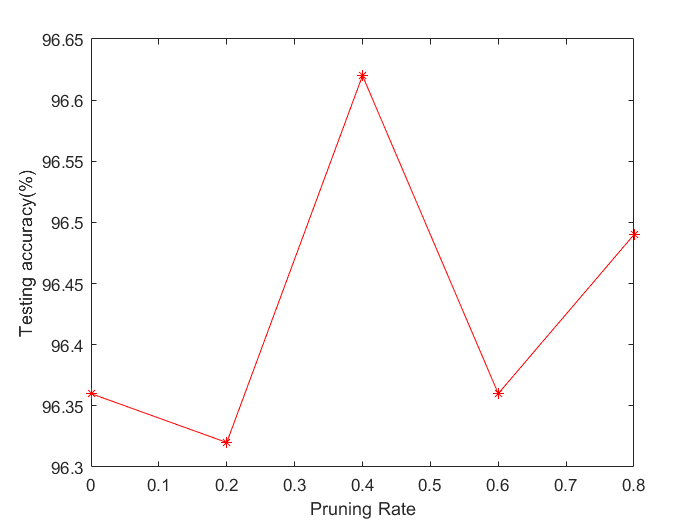}
}
\quad
\subfigure[waveform-noise]{
\includegraphics[width=5.5cm]{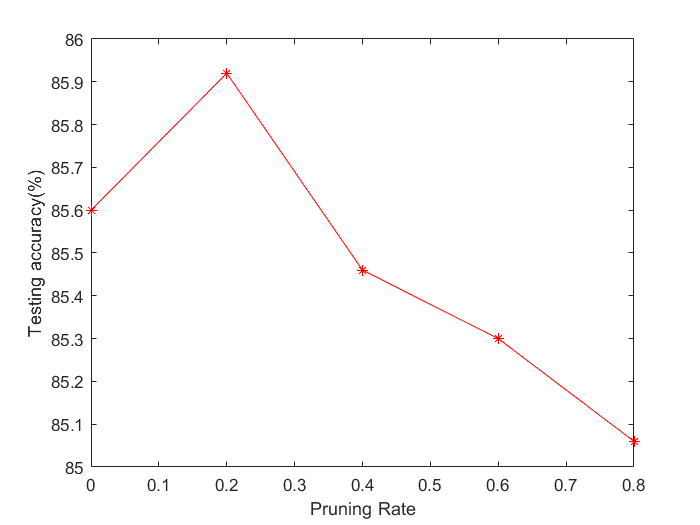}
}
\caption{The testing accuracy of WPedRVFL network with different pruning rate $p$ on 4 tabular datasets.}
\label{figure:pruning rate}
\end{figure}

In addition to that, we present WPedRVFL's performance of changing the pruning rate $p$ in Fig. \ref{figure:pruning rate}. For all 4 tabular datasets, the pruning method can always help us get better accuracy. The four line charts show a similar pattern: At the beginning, the testing accuracy increases with the pruning rate. After reaching a peak value, the accuracy drops and becomes even worse than the initial value. So same as the weight $\omega_r$, we need to set different pruning rates $p$ for different cases using the validation set.

\subsection{Training Time Comparison}

Another advantage of randomized neural networks is their training time. Since they do not use back-propagation to train the hidden parameters, the training time of these RVFL variants can be much less than the deep learning models.

In this section, we select two representatives from two groups of the classifiers. Among the back-propagation based deep neural networks, we choose SNN since it has the highest accuracy on the benchmark datasets. Meanwhile, the most competitive classification model WPedRVFL is selected from the randomized neural networks. We present their training time (without hyperparameter tuning) on 5 selected datasets in Table \ref{Tab:time}.

\begin{table}[h!]
    \centering
    \begin{threeparttable}
    \caption{Training Time comparisons between SNN and WPedRVFL on spambase dataset (s)} 
    \label{Tab:time}
    \begin{tabular}{l c c}\toprule
        Dataset & SNN & WPedRVFL\\ \midrule 
        arrhythmia & 4.4731 & 0.6504 \\
        contrac & 2.8705 & 0.4898 \\
        musk\_2 & 7.0369 & 0.9814 \\
        spambase & 8.6586 & 1.3752 \\
        waveform & 6.7168 & 0.8336\\
        \bottomrule
        \end{tabular}
        \begin{tablenotes}[flushleft]
            \small
                \item $^*$Experiment environment: Intel(R) Xeon(R) CPU E5-2620; nVIDIA GeForce GTX-1080.
        \end{tablenotes}
    \end{threeparttable}    
\end{table}

\section{Conclusion}
In this paper, we first introduce batch normalization to the edRVFL network for re-normalizing the hidden features. Then, we propose a weighted version for edRVFL network. A weight matrix is used to allocate different weights to different samples. The weight matrix changes according to the samples' predictions in the previous layers. This method can make sure that each hidden layer in the network has different biases for each sample and increase the ensemble classification accuracy. Moreover, we propose another new variant of edRVFL with the pruning method. Instead of pruning neurons after the training process, we cut off the inferior neurons according to their importance for classification when we are training the model. This method can prevent the propagation of detrimental features and increase the classification accuracy in deeper layers.
Then, the combination of these two methods called Weighting and Pruning based Ensemble Deep Random Vector Functional Link Network is proposed. It takes advantage of both WedRVFL and PedRVFL and performs better overall. For evaluating the performance of our new proposed methods, we compare them, with other 8 classifiers on 24 UCI benchmark datasets. The experimental results show the superiority of our new methods on tabular classification tasks. In particular, WPedRVFL is the most competitive one among all
the 12 classifiers. In addition to that, we investigate the effects of setting different weight and pruning rate values. These results illustrate how the weighting and pruning schemes can help to improve the classification results. At last, we compare the training time between our proposed model with a back-propagation based deep neural network. In our future work, we will develop methods to select only a few output layers with the highest classification accuracy to perform the final classification.

\newpage 
\bibliography{references.bib}



\end{document}